\documentclass[runningheads]{llncs}
\usepackage[T1]{fontenc}
\usepackage{graphicx}
\usepackage{subcaption}
\usepackage{xcolor}
\usepackage{soul}

\usepackage[toc,page]{appendix}
\begin{document}
\title{CaRTS: Causality-driven Robot Tool Segmentation from Vision and Kinematics Data}
\titlerunning{Causality-driven Robot Tool Segmentation}
%
\author{
Hao Ding$^1$ \and
Jintan Zhang$^1$ \and
Peter Kazanzides$^1$ \and \\
Jie Ying Wu$^2$ \and 
Mathias Unberath$^1$}


\authorrunning{H. Ding et al.}
%
\institute{$^1$Department of Computer Science, Johns Hopkins University
\\ $^2$Department of Computer Science, Vanderbilt University\\
\email{hding15@jhu.edu, unberath@jhu.edu}}


\maketitle              
\begin{abstract} Vision-based segmentation of the robotic tool during robot-assisted surgery enables downstream applications, such as augmented reality feedback, while allowing for inaccuracies in robot kinematics. With the introduction of deep learning, many methods were presented to solve instrument segmentation directly and solely from images. While these approaches made remarkable progress on benchmark datasets, fundamental challenges pertaining to their robustness remain. We present CaRTS, a causality-driven robot tool segmentation algorithm, that is designed based on a complementary causal model of the robot tool segmentation task. Rather than directly inferring segmentation masks from observed images, CaRTS iteratively aligns tool models with image observations by updating the initially incorrect robot kinematic parameters through forward kinematics and differentiable rendering to optimize image feature similarity end-to-end. We benchmark CaRTS with competing techniques on both synthetic as well as real data from the dVRK, generated in precisely controlled scenarios to allow for counterfactual synthesis. On training-domain test data, CaRTS achieves a Dice score of 93.4 that is preserved well (Dice score of 91.8) when tested on counterfactually altered test data, exhibiting low brightness, smoke, blood, and altered background patterns. This compares favorably to Dice scores of 95.0 and 86.7, respectively, of the SOTA image-based method. 
Future work will involve accelerating CaRTS to achieve video framerate and estimating the impact occlusion has in practice. Despite these limitations, our results are promising: In addition to achieving high segmentation accuracy, CaRTS provides estimates of the true robot kinematics, which may benefit applications such as force estimation. Code is available at: \url{https://github.com/hding2455/CaRTS} 

\end{abstract}
\section{Introduction} \label{introduction}
\begin{figure}[t]
\centering
\includegraphics[width=0.9\textwidth]{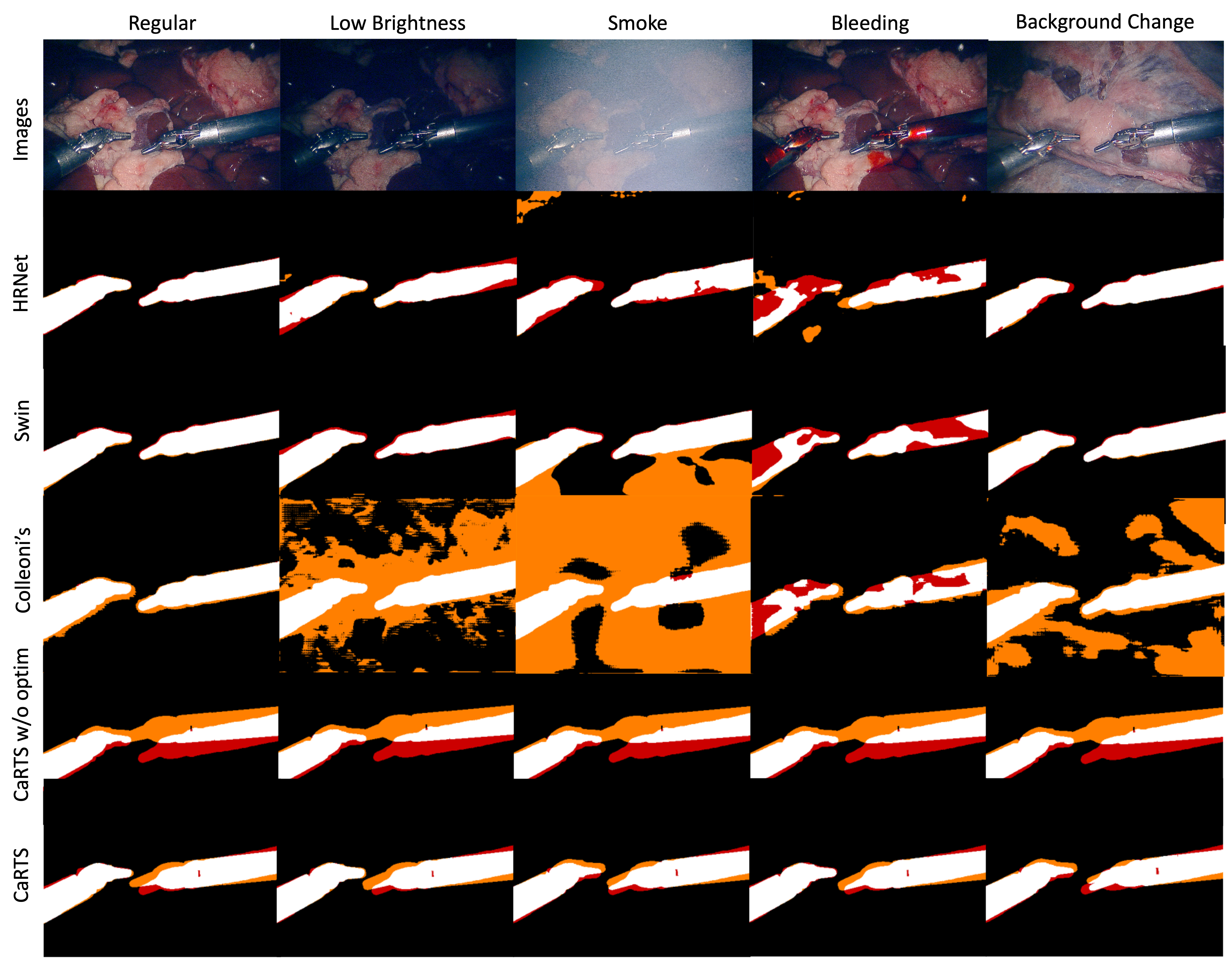}
\caption{Comparisons between HRNet, Swin Transformer, Colleoni's method, and CaRTS with/without optimization on images with counterfactual surgical environment. White/black: True positives/negatives; Orange/Red: False positives/negatives.}
\label{eye_candy}
\end{figure}

With the increasing prevalence of surgical robots, vision-based robot tool segmentation has become an important area of research~\cite{ToolNet,ITP,AIS,DRLISRS,rtis,SICF,one-to-many,su2018real,da2019self,UCL}. Image-based tool segmentation and tracking is considered important for downstream tasks, such as augmented reality feedback, because it provides tolerance to inaccurate robot kinematics. Deep learning techniques for tool segmentation now achieve respectable performance on benchmark dataset, but unfortunately, this performance is not generally preserved when imaging conditions deteriorate or deviate from the conditions of the training data~\cite{RobustnessDL}. Indeed, when encountering conditions of the surgical environment not seen in training, such as smoke and blood as shown in Fig.~\ref{eye_candy}, the performance of contemporary algorithms that seek to infer segmentation masks directly from images deteriorates dramatically. While some of this deterioration can likely be avoided using modern robustness techniques~\cite{RobustnessDL}, we believe that a lot more can be gained by revisiting how the tool segmentation problem is framed.  

\begin{figure}[t]
\begin{subfigure}[b]{0.43\textwidth}
         \centering
         \includegraphics[width=0.9\textwidth]{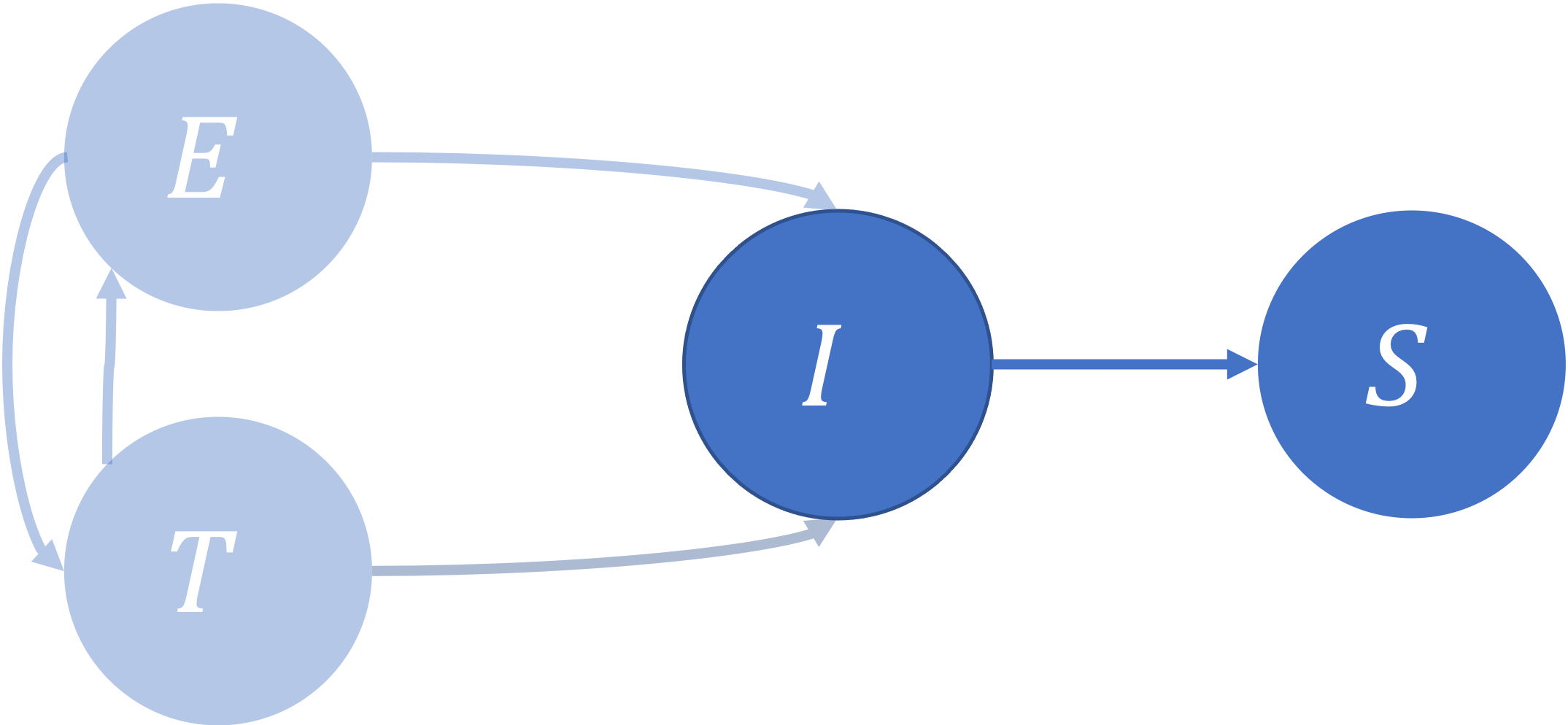}
         \caption{The contemporary causal model}
         \label{contemporary_causal}
     \end{subfigure}
 \hfill
 \begin{subfigure}[b]{0.56\textwidth}
         \centering
         \includegraphics[width=0.9\textwidth]{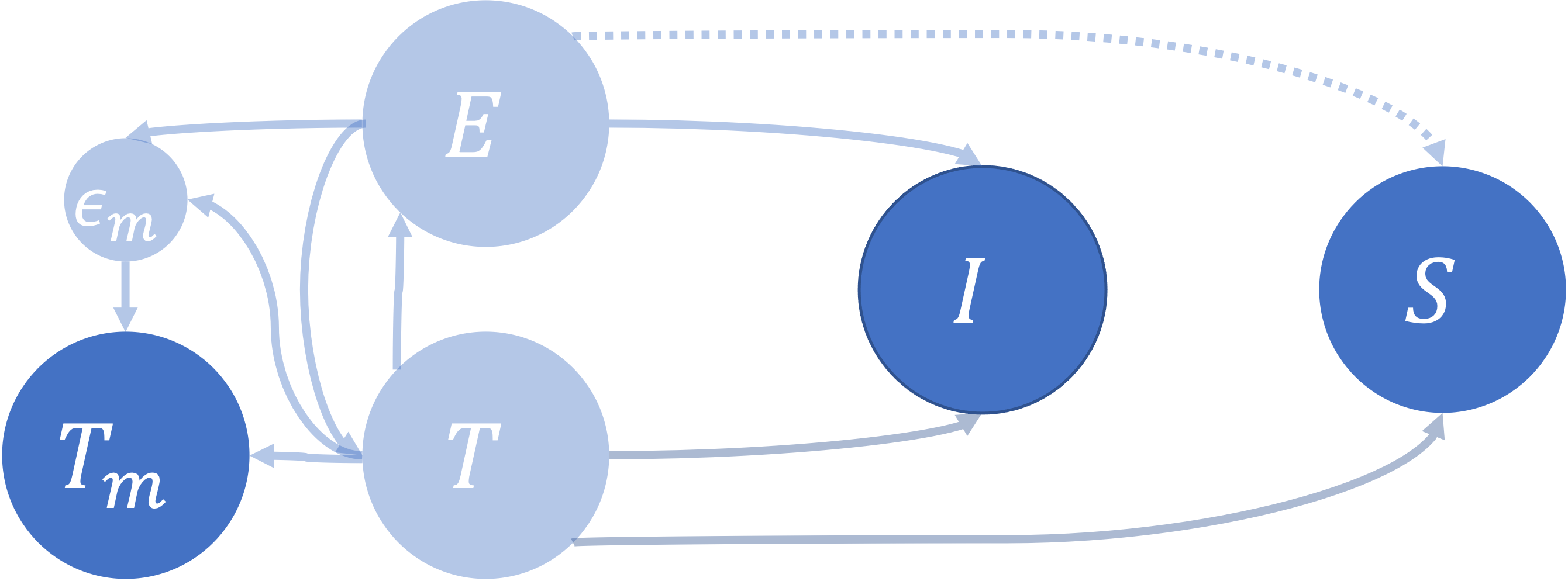}
         \caption{Our causal model}
         \label{our_causal}
     \end{subfigure}
\caption{Illustration of two causal models for robot tool segmentation. Solid lines mean direct causal effect. Light blue represent unobserved factors; dark blue nodes represent observed factors. We note $\mathbf{I}$ for images, $\mathbf{E}$ for environment, $\mathbf{S}$ for segmentation, $\mathbf{T}$ for true robot kinematics and camera poses, $\mathbf{T_m}$ for measured robot kinematics and camera poses, and $\mathbf{\epsilon_m}$ for measurement error.}
\label{image_causal_view}

\end{figure}

In robot tool segmentation, although a rich variety of information is available from the system (e.\,g. kinematics), contemporary approaches to tool segmentation often neglect the complex causal relationship between the system, the surgical environment, and the segmentation maps. These methods posit a direct causal relationship between the observed images and the segmentation maps, following a causal model of segmentation as shown in Fig.~\ref{contemporary_causal}. It follows that the segmentation $\mathbf{S}$ is entirely determined by the image $\mathbf{I}$. Information from robot tools and cameras poses (jointly represented by $\mathbf{T}$), and the environment $\mathbf{E}$ are neither observed nor considered. We refer to this model of determining segmentation maps solely from images as the contemporary model; a causal relation that has previously been suggested for for brain tumor segmentation~\cite{Causality_matters_in_medical_imaging}. 

Under this causal model, there is no confounding bias and one can simply fit $p(\mathbf{S} \mid \mathbf{I})$ which identifies the probability $p(\mathbf{S}(\mathbf{I}))$ for the counterfactual $\mathbf{S}(\mathbf{I})$.
However, these models must learn to generalize solely from the image domain. This makes them vulnerable to domain shifts such as those introduced by lighting changes or smoke during surgery~\cite{DatasetShiftinMachineLearning,domain_survey}. 

We present an alternative causal model of robot tool segmentation, shown in Fig.~\ref{our_causal}, and an end-to-end method designed based on this model, which we refer to as CaRTS: causality-driven robot tool segmentation. In this model, image $\mathbf{I}$ has no direct causal effect on the segmentation $\mathbf{S}$. Instead, both of them are directly determined by the robot kinematics and camera poses $\mathbf{T}$, and the environment $\mathbf{E}$. Interactions between tools and environment are represented by direct causal effects between $\mathbf{E}$ and $\mathbf{T}$. In this causal model, we focus on counterfactual $\mathbf{S}(\mathbf{T})$ instead of $\mathbf{S}(\mathbf{I})$.

Since our models of the robot and the environment are imperfect, there are two main sources of errors for algorithms designed based on this causal model - confounding bias and estimation error. The confounding bias arises from the environment. The causal effect of the environment on the segmentation consists of occlusion. Here, we assume occlusion does not affect segmentation. Under this assumption, the causal relationship between the environment and the segmentation map can be ignored, denoted by the dashed line in Fig.~\ref{our_causal}. The second source of error, estimation error, is caused by the fact that surgical robots are designed to be compliant. Therefore, there is significant error in tool pose estimation based on joint positions. This causes us to observe $\mathbf{T_m}$, which contains measurement errors $\mathbf{\epsilon_{m}}$, instead of the ground truth values $\mathbf{T}$. Our goal is then to estimate $\mathbf{\hat{S}}$ from $\mathbf{T_m}$. Since $\mathbf{I}$ is not affected by $\mathbf{\epsilon_m}$, we can use the information provided by $\mathbf{I}$ to estimate $\mathbf{T}$ based on $\mathbf{T_m}$. 
More specifically, CaRTS uses differentiable rendering given robot and camera information to model $p(\mathbf{S} | \mathbf{T})$ that identifies $p(\mathbf{S}(\mathbf{T}))$. To account for measurement error, CaRTS iteratively estimates $\mathbf{\epsilon_m}$ and infers $\mathbf{T}$ by optimizing a feature-level cosine similarity between observed images $\mathbf{I}$ and images rendered from $\mathbf{T_m}$. 


CaRTS has the advantage that the image is solely used to correct for $\mathbf{T_m}$, and feature extraction reduces reliance on specific pixel representations. As we will show, this makes CaRTS less sensitive to domain shifts in the image domain. Since image appearance depends on the environment, which potentially has high degrees of variability, achieving robustness to image domain shift is challenging. In CaRTS, the segmentation map is conditioned on robot kinematics and camera poses, which by design limits the space of possible segmentation. Variability in these parameters is restricted and changes in kinematics and camera poses can often be accounted for by small adjustments.

The main contributions of this paper are: (1) a complementary causal model for robot tool segmentation that is robust to image domain shifts, and (2) an algorithm, CaRTS, based on (1) that generalizes well to new domains.

\section{Related Work}

\paragraph{\textbf{Robot Tool Segmentation:}} Vision-based segmentation is a mature area of research and many deep learning-based methods, including~\cite{unet,FCN,deeplabv3+,maskrcnn,HTC,DSC,liu2021Swin,WangSCJDZLMTWLX19}, exist. For robot tool segmentation, many techniques follow the same paradigm and solely rely on vision data to achieve scene segmentation, e.\,g.,~\cite{ToolNet,DRLISRS,AIS,ITP,rtis,one-to-many}. More recently and related to this work, there have been efforts to combine vision-based techniques with other sources of information, such as robot kinematics, which is hypothesized to improve segmentation performance~\cite{SICF} or robustness~\cite{UCL}. To this end, Su et al.~\cite{su2018real} use frequency domain shape matching to align a color filter- and kinematic projection-based robot tool mask. Similarly, da Costa Rocha et al.~\cite{da2019self} optimize for a base-frame transformation using a grab-cut image similarity function to create well annotated data in a self-supervised fashion to later train a purely image-based model.  Colleoni et al.~\cite{UCL} train a post-processing neural network for image-based refinement of a kinematic-based projection mask.  While these recent methods use geometric data in addition to images, none of the previous methods enables direct estimation of robot kinematic parameters, including joint angles, base frame and camera transformations, in an end-to-end fashion.

\paragraph{\textbf{Geometric information for robot vision:}} Geometric information (e.\,g. depth, pose) has been explored and previously been used in robot vision. Some methods~\cite{LiYWCTVU2021temporal,LongLYNTUD21edssr,LiLDDCTU21sttr,AllanOHKS18,YeZGY16rt3dtracking,GodardAFB19,GuoYYWL19} are proposed to acquire geometric information from vision, other methods~\cite{UCL,SICF,su2018real,da2019self,HazirbasMDC16fusenet,couprie2013indoor} use geometric information to enhance downstream vision tasks. Our method leverages both directions by optimizing the measured robot and camera parameters from vision and enhancing the segmentation of the image via projection of the robot.


\paragraph{\textbf{Causality in Medical Image Analysis:}} 
Recently, causality receives increasing attention in the context of medical image analysis~\cite{Causality_matters_in_medical_imaging}. Reinhold et al.~\cite{causalMRI} use causal ideas to generate MRI images with a Deep Structural Causal Model~\cite{DSCM}. Lenis et al.~\cite{Domainaware} use the concept of counterfactuals to analyze and interpret medical image classifiers. Lecca~\cite{Perspectives} provides perspectives on the challenges of machine learning algorithms for causal inference in biological networks. Further, causality-inspired methods~\cite{stable,RLIC,causalDomain,cvrnn,causalSemantic} exist for domain generalization~\cite{domain_survey}. These techniques focus on feature representation learning directly from images, i.\,e., the contemporary causal model of segmentation. Different from these methods, our approach frames the robot tool segmentation task using an alternative causal model, and leverages this model to derive a different approach to segmentation that alters the domain shift problem.

\section{CaRTS: Causality-driven Robot Tool Segmentation}
We propose the CaRTS architecture as an algorithm designed based on our causal model. CaRTS models $p(\mathbf{S} | \mathbf{T})$, which identifies $p(\mathbf{S(T)})$ through differentiable rendering, to estimate segmentation maps from robot kinematic parameters. CaRTS estimates the measurement error $\mathbf{\epsilon_m}$ between true and observed robot kinematics  ($\mathbf{T}$ and $\mathbf{T_m}$, respectively) by maximizing a cosine similarity between the semantic features of the observed and rendered image extracted by a U-Net. 

\begin{figure}[t]
    \centering
    \includegraphics[width=0.9\textwidth]{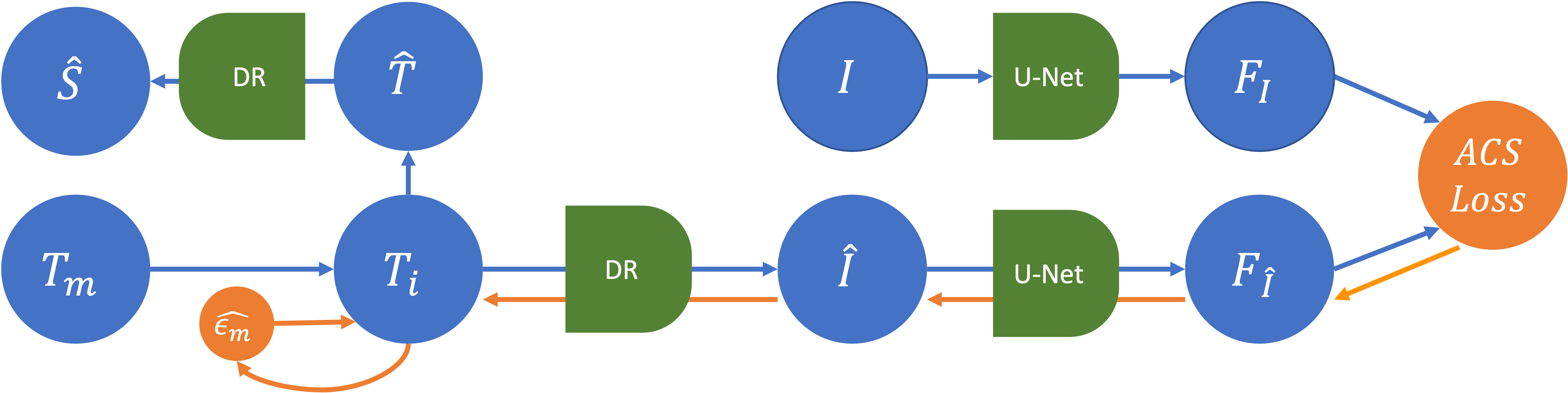}
    \caption{Illustration of the overall CaRTS architecture. 
    We iteratively optimize ACL Loss w.r.t $\mathbf{T_i}$ to estimate robot parameters $\mathbf{\hat{T}}$ then generate estimated segmentation $\mathbf{\hat{S}}$.
    solid blue lines in this figure represent information flow and solid orange lines represent gradient flow. DR stands for differentiable rendering.}
    \label{CaRTS}
\end{figure}

\paragraph{\textbf{Overall Architecture:}} The overall architecture of CaRTS is shown in Fig.~\ref{CaRTS}. We observe $\mathbf{T_m}$ and $\mathbf{I}$, the measured kinematic parameters and an image, as input to our model. $\mathbf{T_i}$ represents the values of robot and camera parameters at iteration $i$ of the optimization, with $\mathbf{T_0}$ being initialized to $\mathbf{T_m}$. 

For each iteration, CaRTS uses differentiable rendering to generate an estimated image $\mathbf{\hat{I}}$ from $\mathbf{T_i}$ and articulated tool models. Then it uses a pre-trained U-Net to extract features $\mathbf{F_I}$ and $\mathbf{{F_{\hat{I}}}}$ from both observed $\mathbf{I}$ and rendered image $\mathbf{\hat{I}}$, respectively. Attentional cosine similarity loss (ACS Loss) between the feature representations is used to evaluate the similarity of both images. Since the whole pipeline is differentiable, we can use backpropagation to directly calculate the gradient to estimate the measurement error $\mathbf{\epsilon_m}$. We note this estimation as $\mathbf{\hat{\epsilon_m}}$. So with gradient descent, we iteratively estimate $\mathbf{T}$ from $\mathbf{T_i}$ by subtracting $\mathbf{\hat{\epsilon_m}}$ multiplied by a step size. We consider $\mathbf{T_i}$ with the lowest loss as our estimated $\mathbf{\hat{T}}$, which is used to estimate the final segmentation $\mathbf{\hat{S}}$.


\paragraph{\textbf{Differentiable Rendering:}}
Unlike traditional rendering techniques, differentiable rendering enables the calculation of gradient w.r.t. the 3D quantities which has been found beneficial for related projective spatial alignment tasks~\cite{ProST}. We use the differentiable rendering function (noted as $f_{DR}$) provided by Pytorch3d~\cite{ravi2020pytorch3d}, implementing Soft Rasterizer~\cite{softRasterizer} and Neural 3D Mesh Renderer~\cite{Neural3D}. For CaRTS, differentiable rendering enables the calculation of gradient w.r.t. the robot parameters used to calculate the final robot meshes and camera parameters.  We calculate forward kinematics (denoted by the function $f_{FK}$)~\cite{fontanelli2017Modelling} for the robot to construct meshes $\mathbf{M}$ for rendering. We render image $\mathbf{\hat{I}}$ and segmentation $\mathbf{\hat{S}}$ using one point light centered behind the camera position $\mathbf{L_d}$. This yields:
\begin{equation}
\mathbf{M} = f_{FK}(\mathbf{DH}, \mathbf{M_B}, \mathbf{F_B}, \mathbf{K})\qquad\qquad\qquad
\mathbf{\hat{I}},\mathbf{\hat{S}} = f_{DR}(\mathbf{C}, \mathbf{M}, \mathbf{L_d})
\label{equ_render}
\end{equation}

\paragraph{\textbf{Neural Network Feature Extraction:}}
%
We train a U-Net~\cite{unet} on a dataset consisting of two types of images: The originally collected images $\mathbf{I}$ from the training set, and hybrid images denoted as $\mathbf{BG_m} + \mathbf{\hat{I}}$. $\mathbf{BG_m}$ is the pixel-wise average background of all originally collected images in the training set. $+$ means overlaying the average background with the robot tool content of the rendered images $\mathbf{\hat{I}}$. Training on hybrid images ensures that feature extraction yields reasonable features from both acquired and rendered images. The average background replaces 0s in the rendered image to avoid potential numerical issues. All hybrid images and the average background can be obtained "for free" from the training set. We optimize the binary cross-entropy loss between the network output and the ground-truth segmentation for both the original and hybrid images to make the network learn semantic feature representations. At inference time, we extract feature maps from the last feature layer for both $\mathbf{I}$ and $\mathbf{BG_m} + \mathbf{\hat{I}}$. We note the feature extraction function as $f_{FE}$. This yields:
\begin{equation}
\mathbf{F_I} = f_{FE}(\mathbf{I})\qquad\qquad\qquad
\mathbf{F_{\hat{I}}} = f_{FE}(\mathbf{\hat{I}} + \mathbf{BG_m})
\label{equ_feature_extra}
\end{equation}
Although our feature extractor is trained for segmentation, other training targets beyond segmentation, such as unsupervised representation learning, might be equally applicable and more desirable.

\paragraph{\textbf{Attentional Cosine Similarity Loss:}}
We design the Attentional Cosine Similarity Loss (ACSLoss) to measure the feature distance between $\mathbf{F_{\hat{I}}}$ and $\mathbf{F_I}$. First, we generate an attention map $\mathbf{Att}$ by dilating the rendered silhouette $\mathbf{\hat{S}}$ of the robot tools. This attention map allows us to focus on the local features around the approximate tool location given by the current kinematics estimate and DR. Then we calculate cosine similarity along the channel dimension of the two feature maps. The final loss is calculated using equation~\ref{equ_ACSLoss}. Higher channel-wise cosine similarity suggests better alignment between feature points:
\begin{equation}
\mathbf{Att} = dilate(\mathbf{\hat{S}})\qquad\qquad
\mathbf{L} = 1 - \frac{1}{w \times h}\sum_{i = 0}^{h - 1}\sum_{j = 0}^{w - 1}sim_{cos}(\mathbf{F_I}, \mathbf{F_{\hat{I}}})_{i,j} \times \mathbf{Att}_{i,j}
\label{equ_ACSLoss}
\end{equation}

The complete objective for optimization is shown in euqation~\ref{equ_optim_obj}. While in this paper we fix all but the kinematic parameters and optimize $f_L$ w.r.t the kinematics input $\mathbf{K}$ to achieve tool segmentation, we note that optimization w.r.t. these variables corresponds to possibly desirable robotic applications. For example, if we fix all variables and optimize $f_L$ w.r.t the camera parameters $\mathbf{C}$ and robot base frame $\mathbf{F_B}$, we can perform hand-eye calibration:
\begin{equation}
\arg\min f_L(\mathbf{DH}, \mathbf{M_B}, \mathbf{F_B}, \mathbf{K}, \mathbf{C}, \mathbf{L_d}, \mathbf{I}, \mathbf{BG_m})
\label{equ_optim_obj}
\end{equation}

\section{Experiments}
We use the da Vinci Research Kit (dVRK)~\cite{dvrk} as the platform for data collection. The system has a stereo endoscope and two patient side manipulators (PSM).
The task is differentiate PSMs from the rest of the endoscope scene through segmentation. We conduct robot tool segmentation experiments on synthetic data generated using AMBF~\cite{ambf} and real-world data collected with the dVRK, respectively. Under this setting, CaRTS perform optimization w.r.t a 6 degree-of-freedom joint space. Implementation details can be found in the released code.

\paragraph{\textbf{Data Collection:}}
For simulated data, we generate 8 sequences (6 for training, 1 for validation, and 1 for testing), by teleoperating the PSMs in the simulator, recording the image, segmentation and kinematics read from the simulator. Each sequence contains 300 images. Robot base frame and camera information are determined before recording. For test data, we add random offsets to the kinematics to simulate measurement error and create another test set consisting of images that are corrupted with simulated smoke.

For real data, we adopt the data collecting strategy from Colleoni et al.~\cite{UCL}, but expand it to allow for counterfactual data generation. We prerecord a kinematics sequence and replay it on different backgrounds. The first replay is on a green screen background to extract ground truth segmentation through color masking. Then, we introduce different tissue backgrounds and set different conditions to counterfactually generate images that differ only in one specific condition - every condition will be referred to as a domain. We choose one domain without corruption for training, validation and test. This regular domain has a specific tissue background, a bright and stable light source, and no smoke or bleeding conditions. Images collected from other domains are only provided for validation and testing. We record 7 sequences for training, 1 for validation, and 1 for testing. Each sequence contains 400 images. To ensure replay accuracy, we keep the instrument inserted during the whole data collection procedure.

\begin{figure}[t]
\centering
     \includegraphics[width=0.9\textwidth]{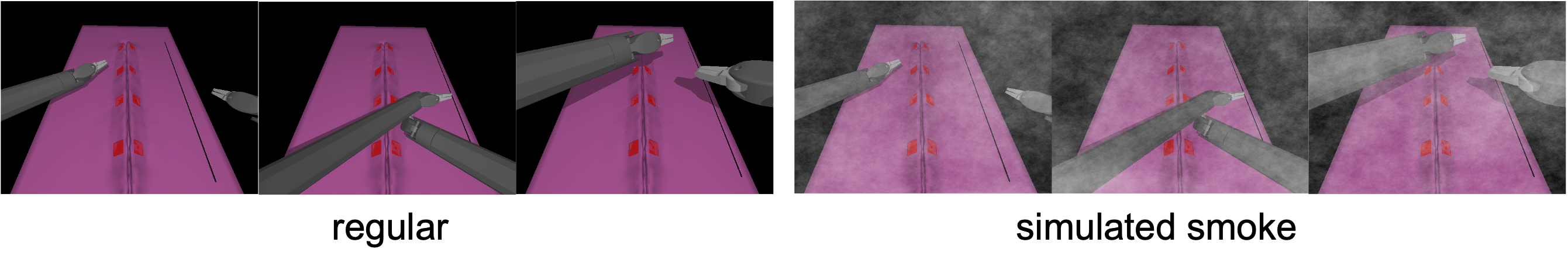}
\caption{Examples of the images in the synthetic dataset.}
\label{image_examples}
\end{figure}

All segmentation maps are generated under no occlusion condition to avoid confounding issues. Fig.~\ref{eye_candy} and Fig.~\ref{image_examples} show examples of the real-world and synthetic datasets respectively.

\paragraph{\textbf{Experiment on Robot Tool Segmentation:}} In this experiment, we compare the tool's Dice score of CaRTS to the HRNet~\cite{WangSCJDZLMTWLX19}, Swin Transformer~\cite{liu2021Swin} trained with simulated smoke augmentation, CaRTS without optimization and method by Colleoni et al.~\cite{UCL} using both kinematics and images.

\begin{table}[t]
\centering
\caption{Robot Tool Segmentation results}\label{tab_real_word}
\resizebox{\columnwidth}{!}{%
\begin{tabular}{|c|c|c|c|c|c|c|c|}
\hline
 & R-Reg  & R-LB & R-Bl & R-Sm & R-BC & S-Reg & S-SS \\
\hline
Colleoni's & $94.9 \pm 2.7$ &$87.0 \pm 4.5$ & $55.0 \pm 5.7$& $59.7 \pm 24.7$& $75.1 \pm 3.6$& $99.8 \pm 0.1$& $41.4 \pm 4.3$ \\
HRNet w Aug& $ 95.2 \pm 2.7$ & $ 86.3 \pm 3.9 $ & $ 56.3 \pm 16.4$ & $ 77.2 \pm 23.6$ & $92.1  \pm 4.6 $ &   - & - \\
Swin Transformer w Aug &  $95.0\pm 5.5$  &  $ 93.0\pm 5.5$ &  $76.5 \pm 9.0$    &  $82.4\pm 17.0$  & $94.8\pm 5.3$ & - & - \\
CaRTS w/o Optim.  & $89.4 \pm 3.9$  & - & - & - & - & $83.4 \pm 6.8$& - \\
CaRTS L1  & $92.7 \pm 4.0$ & $89.8 \pm 3.9$& $89.4 \pm 3.9$ & $89.8 \pm 3.9$ & $89.5 \pm 3.9$ & $90.5 \pm 10.0$ & $83.7 \pm 6.9$\\
CaRTS ACS &  $93.4 \pm 3.0$ & $92.4 \pm 3.1$ & $90.8 \pm 4.4$& $91.6 \pm 4.7$ & $92.3 \pm 4.8$ & $96.6 \pm 2.7$& $96.5 \pm 2.4$\\
\hline
\end{tabular}%
}
\end{table}

The results are summarized in Tab.~\ref{tab_real_word}. Each column represents a test domain. "R-" and "S-" represent real-world data and synthetic data, respectively. Reg, LB, Bl, Sm, BC, and SS represent regular, low brightness, bleeding, smoke, background change, and simulated smoke respectively. As the table shows, HRNet and Swin Transformer perform well on the regular domain on both real and synthetic data. However, their performance deteriorates notably when the testing domain changes. Colleoni et al.’s architecture performs similarly well on the regular domain, and compared to HRNet and Swin Transformer and still suffers from performance deterioration in unseen testing domains. CaRTS achieves similarly high performance irrespective of test domain.

\paragraph{\textbf{Ablation Study:}}
We perform ablation studies on the validation set to explore the impact of the similarity function, iteration numbers, and magnitude of kinematics errors on segmentation performance. Fully-tabled results for the latter two ablation studies are not shown here due to the space limit. 

As shown in Tab.~\ref{tab_real_word}, we find that optimizing ACSLoss on features better utilizes the rich information extracted by the U-Net compared to optimizing a pixel-level smooth-L1 loss on predictions, especially on unseen domains. The ablation study on the number of iterations indicates that even though optimization on different test domains performs similarly after a single step. Only some domains continue to improve with larger numbers of iterations (iteration > 30). It takes from 30 to 50 iterations on average for the optimization to converge on different domains.
The ablation study on kinematic error magnitude indicates that final segmentation performance depends more on initial mismatch in the image space rather than in the joint space. When initial Dice score is less than a threshold (around 75 in our setting), CaRTS's performance starts dropping.

\section{Conclusion}

While CaRTS achieved robustness across different domains, its current implementation has some limitations. First, the optimization for $\mathbf{T}$ typically requires over 30 iterations to converge and is thus not yet suitable for real-time use. Including temporal information may reduce the number of iterations necessary or reduce the rendering time at each iteration. Second, from a theoretical perspective, CaRTS does not allow for occlusions. To address occlusion, future work could evaluate the importance of this theoretical limitation in practice, and explore disentangling environment features from images and use adjustment equation to predict segmentation.

The benchmark results on both synthetic and real dataset generated in precisely controlled scenarios show that, unlike segmentation algorithms based solely on images, CaRTS performs robustly when the test data is counterfactually altered. In addition to high segmentation accuracy and robustness, CaRTS estimates accurate kinematics of the robot, which may find downstream application in force estimation or related tasks.

\paragraph{\textbf{Acknowledgement:}} This research is supported by a collaborative research agreement with the MultiScale Medical Robotics Center at The Chinese University of Hong Kong.

\bibliographystyle{splncs04}
\bibliography{main}

\newpage
\appendix
\title{Supplementary Materials}
\author{
Hao Ding$^1$ \and
Jintan Zhang$^1$ \and
Peter Kazanzides$^1$ \and \\
Jie Ying Wu$^2$ \and 
Mathias Unberath$^1$}


\authorrunning{H. Ding et al.}
%
\institute{$^1$Department of Computer Science, Johns Hopkins University
\\ $^2$Department of Computer Science, Vanderbilt University\\
\email{hding15@jhu.edu, unberath@jhu.edu}}
\maketitle
\section{Robot Setup} 

\begin{figure}[h]
\centering
     \includegraphics[width=0.9\textwidth]{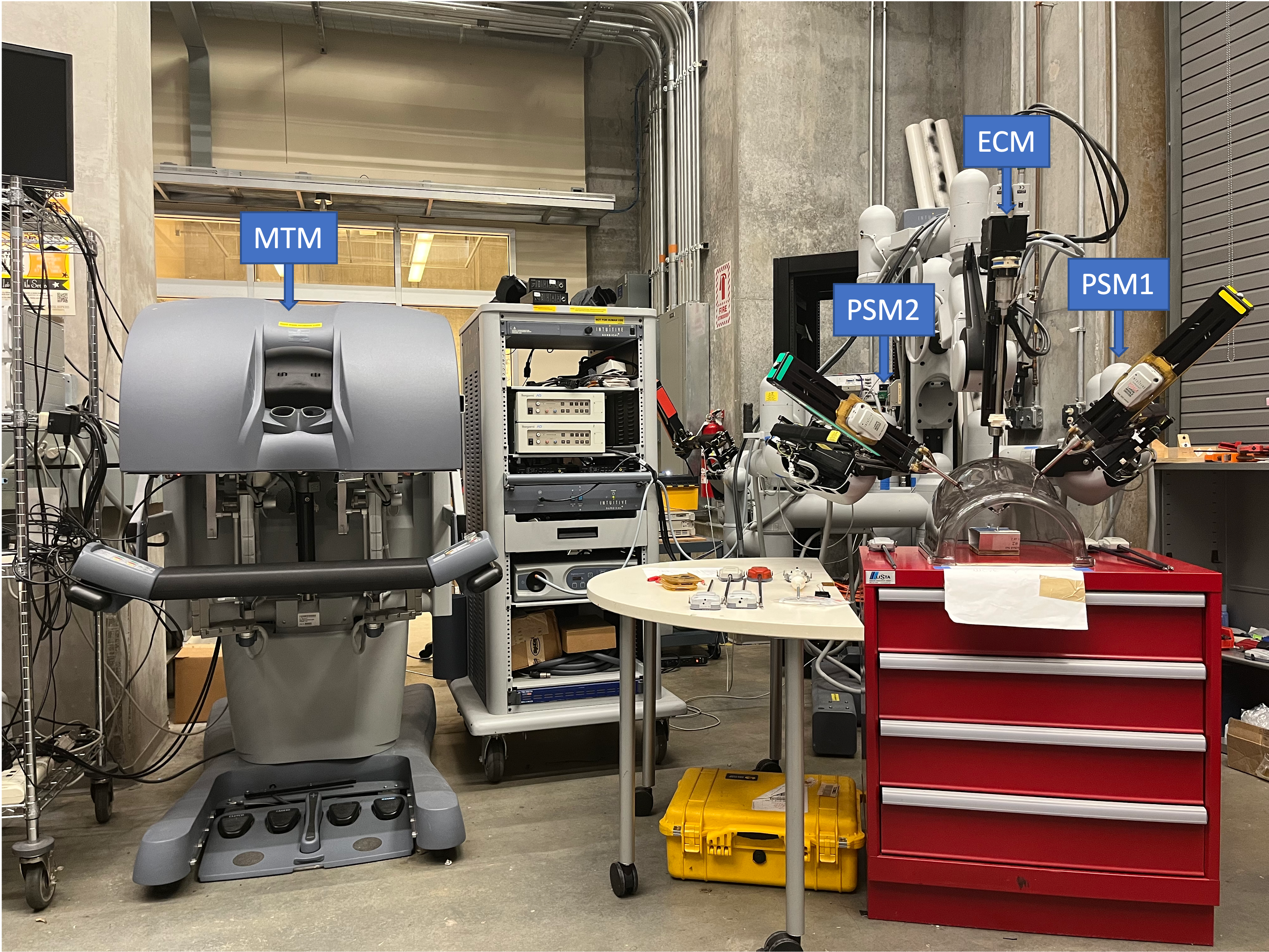}
\caption{Illustration of Robot Setup. Two PSMs (Patient Side Manipulators) are used as robot tool. ECM (Endoscopic Camera Manipulator) is fixed for image capture. MTM (Master Tool Manipulators) is used to manually generate kinematics sequences. The dVRK (da Vinci Research Kit) runs on a PC connected to the whole system to record kinematics, manipulate robots and collect images.}
\label{robot_set_up}
\end{figure}

\section{Ablation Study Results}
\begin{table}
\centering
\caption{Ablation study on the different iteration numbers}\label{tab_iteration}
\par\rule{0pt}{3mm}
\resizebox{\columnwidth}{!}{%
\begin{tabular}{|c|c|c|c|c|c|c|c|}
\hline
 & U-Net DICE & 1 & 10 & 20  & 30& 50 & 100 \\
\hline
R-Reg & $95.0 \pm 2.7$ & $90.4 \pm 3.8$ & $92.6 \pm 3.3$ & $93.1 \pm 3.3$ & $93.2 \pm 3.1$ & $93.5 \pm 3.1$ &  $93.6 \pm 2.9$\\

R-LB & $84.1 \pm 3.5$ & $90.2 \pm 3.7$ & $91.8 \pm 3.4$ & $92.0 \pm 3.1$ & $92.2 \pm 3.0$ & $92.5 \pm 3.0$ &   $92.7 \pm 2.8$\\

R-BC & $76.8 \pm 2.9$ & $90.2 \pm 3.7$ & $91.9 \pm 3.2$ & $92.0 \pm 3.1$ & $92.1 \pm 3.0$ & $92.4 \pm 3.0$ &  $92.4 \pm 3.1$\\

R-Sm & $59.7 \pm 24.7$ & $90.0 \pm 3.9$ & $91.4 \pm 4.3$ & $91.5 \pm 4.6$ & $91.5 \pm 4.8$ & $91.5 \pm 5.2$ &  $91.4 \pm 5.5$\\

R-Bl & $30.6 \pm 13.9$ & $90.2 \pm 3.8$ & $91.3 \pm 3.9$ & $91.1 \pm 4.0$ & $91.1 \pm 4.3$ & $91.1 \pm 4.3$ &  $91.0 \pm 4.5$\\
\hline
\end{tabular}}
\par\rule{0pt}{3mm}
\caption{Ablation study on the different kinematics error. }\label{tab_error}
\resizebox{\columnwidth}{!}{%
\begin{tabular}{|c|c|c|c|c|c|c|c|}
\hline
 Error/Metrics & $0^\circ$ / Dice & $1^\circ$ / Dice & $2^\circ$ / Dice & $3^\circ$ / Dice  & $1^\circ$ / MAE& $2^\circ$ / MAE & $3^\circ$ / MAE \\
\hline
R-Reg & $93.4 \pm 3.1$ & $92.2 \pm 4.3$ & $86.1 \pm 9.1$ & $76.2 \pm 14.0$ & $(0.6 \pm 0.4)^\circ$ & $(1.4 \pm 0.6)^\circ$ &  $(2.6 \pm 0.6)^\circ$\\

S-Reg & $98.1 \pm 0.7$ & $97.9 \pm 1.3$ & $95.3 \pm 4.1$ & $90.3 \pm 5.2$ & $(0.1 \pm 0.1)^\circ$ & $(0.4 \pm 0.4)^\circ$ &  $(1.3 \pm 0.4)^\circ$\\
\hline
\end{tabular}}
\end{table}

\begin{figure}
\centering
\begin{subfigure}[b]{0.49\textwidth}
         
         \includegraphics[width=\textwidth]{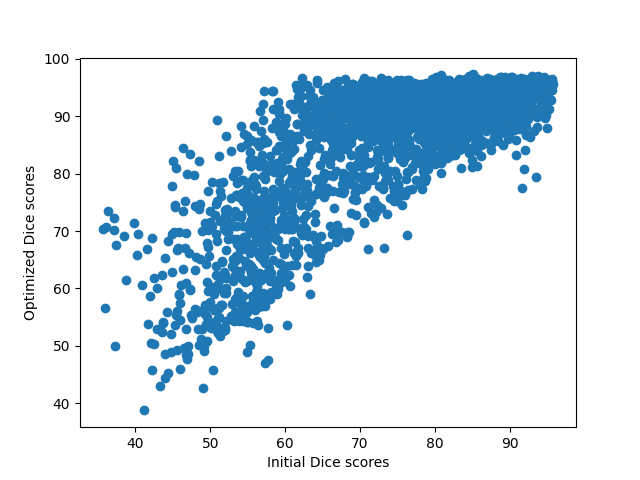}
         
         \caption{\centering{X: initial Dice, Y: optimized Dice test domain: Real-Regular}}
     \end{subfigure}
 \hfill
 \begin{subfigure}[b]{0.49\textwidth}
         \includegraphics[width=\textwidth]{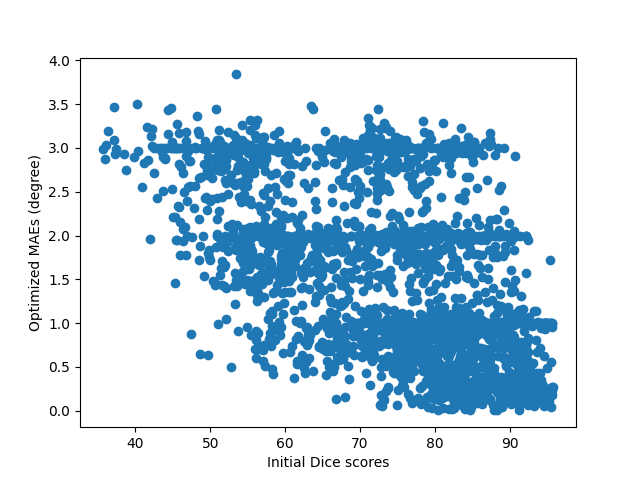}
         \caption{\centering{X: initial Dice, Y: optimized MAE test domain: Real-Regular}}
     \end{subfigure}
\begin{subfigure}[b]{0.49\textwidth}
     \includegraphics[width=\textwidth]{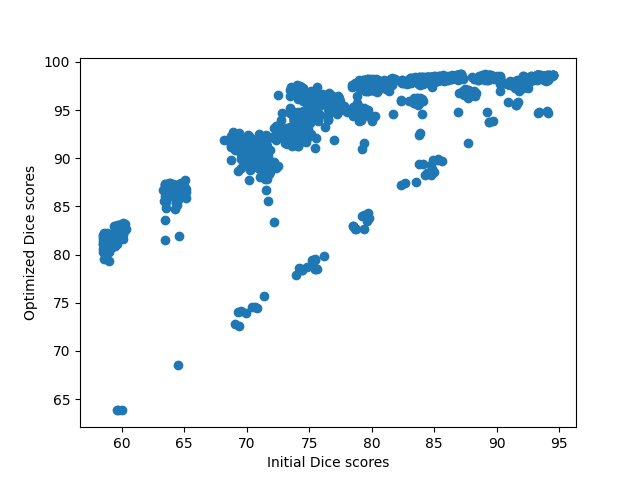}
         \caption{\centering{X: initial Dice, Y: optimized Dice test domain: Synthetic-Regular}}
     \end{subfigure}
 \hfill
 \begin{subfigure}[b]{0.49\textwidth}
         \includegraphics[width=\textwidth]{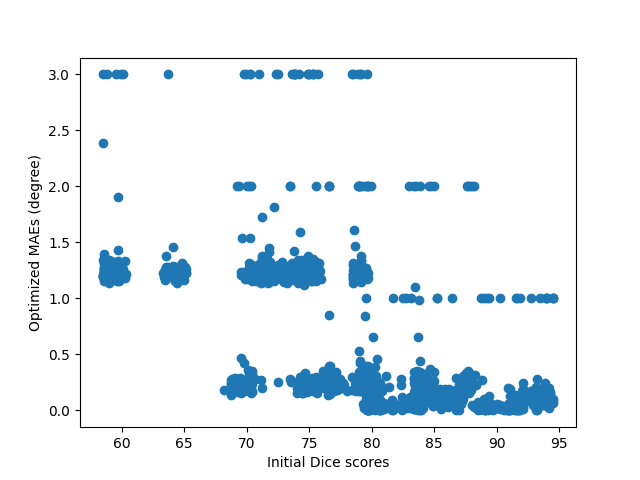}
         \caption{\centering{X: initial Dice, Y: optimized MAE test domain: Synthetic-Regular}}
     \end{subfigure}
\caption{Illustration of the distribution of the test samples in ablation study on the different kinematics error. The final optimized MAEs and Dice scores shows high correlations with the initial Dice scores}
\end{figure}

\end{document}